\documentclass[conference,dvipsnames]{IEEEtran}
\usepackage{cite}
\usepackage{amsmath,amssymb,amsfonts,amsthm}
\usepackage{algorithmic}
\usepackage{graphicx}
\usepackage{textcomp}
\usepackage{xcolor}
\def\BibTeX{{\rm B\kern-.05em{\sc i\kern-.025em b}\kern-.08em
    T\kern-.1667em\lower.7ex\hbox{E}\kern-.125emX}}
\usepackage[english]{babel}
\usepackage[letterpaper,top=2cm,bottom=2cm,left=3cm,right=3cm,marginparwidth=1.75cm]{geometry}

\usepackage[colorlinks=true, allcolors=blue]{hyperref}

\usepackage[utf8]{inputenc} % allow utf-8 input
\usepackage[T1]{fontenc}    % use 8-bit T1 fonts
\usepackage{url}            % simple URL typesetting
\usepackage{booktabs}       % professional-quality tables
\usepackage{nicefrac}       % compact symbols for 1/2, etc.
\usepackage{microtype}      % microtypography
\usepackage{todonotes,setspace}

\newcounter{mycomment}
\newcommand{\mycomment}[2][]{% 4 rotating colors for overlaps
   \refstepcounter{mycomment}%
   \ifcase\numexpr \themycomment- 4*((\themycomment+3)/4 -1) \relax
   {%
      \setstretch{0.7}%spacing
      \todo[color=SeaGreen,size=\scriptsize]{%
         \textbf{[\uppercase{#1}\themycomment]:} #2}%
   }
   \or
   {%
      \setstretch{0.7}%spacing
      \todo[color=Salmon,size=\scriptsize]{%
         \textbf{[\uppercase{#1}\themycomment]:} #2}%
   }
   \or
   {%
      \setstretch{0.7}%spacing
      \todo[color=Thistle,size=\scriptsize]{%
         \textbf{[\uppercase{#1}\themycomment]:} #2}%
   }
   \else
   {%
      \setstretch{0.7}%spacing
      \todo[color=Tan,size=\scriptsize]{%
         \textbf{[\uppercase{#1}\themycomment]:} #2}%
   }
   \fi
}

\begin{document}

\title{Stochastic Neuromorphic Circuits for Solving MAXCUT
% \thanks{}
}

\makeatletter
\newcommand{\linebreakand}{%
  \end{@IEEEauthorhalign}
  \hfill\mbox{}\par
  \mbox{}\hfill\begin{@IEEEauthorhalign}
}
\makeatother

\author{\IEEEauthorblockN{Bradley H. Theilman}
\IEEEauthorblockA{\textit{Neural Exploration and Research Lab} \\
\textit{Sandia National Laboratories}\\
Albuquerque, New Mexico\\
bhtheil@sandia.gov}
\and
\IEEEauthorblockN{Yipu Wang}
\IEEEauthorblockA{\textit{Discrete Math and Optimization}\\
\textit{Sandia National Laboratories}\\
Albuquerque, New Mexico\\
yipwang@sandia.gov}
\linebreakand
\IEEEauthorblockN{Ojas Parekh}
\IEEEauthorblockA{\textit{Discrete Math and Optimization} \\
\textit{Sandia National Laboratories} \\
Albuquerque, New Mexico \\
odparek@sandia.gov}
\and
\IEEEauthorblockN{William Severa}
\IEEEauthorblockA{\textit{Neural Exploration and Research Lab}\\
\textit{Sandia National Laboratories}\\
Albuquerque, New Mexico\\
wmsever@sandia.gov}
\linebreakand
\IEEEauthorblockN{J. Darby Smith}
\IEEEauthorblockA{\textit{Neural Exploration and Research Lab} \\
\textit{Sandia National Laboratories}\\
Albuquerque, New Mexico \\
jsmit16@sandia.gov}
\and
\IEEEauthorblockN{James B. Aimone}
\IEEEauthorblockA{\textit{Neural Exploration and Research Lab} \\
\textit{Sandia National Laboratories}\\
Albuquerque, New Mexico \\
jbaimon@sandia.gov}
}

\maketitle

\begin{abstract}
Finding the maximum cut of a graph (MAXCUT) is a classic optimization problem that has motivated parallel algorithm development.  While approximate algorithms to MAXCUT offer attractive theoretical guarantees and demonstrate compelling empirical performance, such approximation approaches can shift the dominant computational cost to the stochastic sampling operations.  Neuromorphic computing, which uses the organizing principles of the nervous system to inspire new parallel computing architectures, offers a possible solution. One ubiquitous feature of natural brains is stochasticity: the individual elements of biological neural networks possess an intrinsic randomness that serves as a resource enabling their unique computational capacities. By designing circuits and algorithms that make use of randomness similarly to natural brains, we hypothesize that the intrinsic randomness in microelectronics devices could be turned into a valuable component of a neuromorphic architecture enabling more efficient computations.  Here, we present neuromorphic circuits that transform the stochastic behavior of a pool of random devices into useful correlations that drive stochastic solutions to MAXCUT.  We show that these circuits perform favorably in comparison to software solvers and argue that this neuromorphic hardware implementation provides a path for scaling advantages. This work demonstrates the utility of combining neuromorphic principles with intrinsic randomness as a computational resource for new computational architectures.  
\end{abstract}

\section{Introduction}

Despite the heavy requirements for noise-free operation placed on the components of conventional computers, random numbers play a crucially important role in many parallel computing problems arising in different scientific domains. Because current random number generation occurs largely in software, the required randomness in these systems is plagued by the same memory-processing bottlenecks that limit ordinary computation. Current work in material science and microelectronics is demonstrating the feasibility of constructing stochastic microelectronic devices with controllable statistics for probabilistic neural computing \cite{misra_probabilistic_2022}.  These devices show scalability properties that forecast the ability to generate random numbers in-situ with the processing elements, bypassing this bottleneck.  

Stochasticity is an inherent property of physical systems, both natural and artificial.  Physical computers are able to approximate ideal computations because much effort has been expended in developing electronic technology that minimizes the influence of universal electronic ``noise.'' As microelectronics get smaller and the scale of our computations get larger, current computational paradigms require even more stringent limits on the influence of this electronic noise, and these limits become severe constraints on the scalability of existing computational architectures.

In contrast, natural brains are examples of highly parallel computational systems that achieve amazingly efficient computational performance in the face of ubiquitous noise. There are on the order of $10^{15}$ synapses in a human brain, and each one is stochastic: its probability of successfully transmitting a signal to a downstream neuron ranges from 0.1 to 0.9~\cite{Koch1999}. Each synapse is activated about once per second on average, so, the brain generates about $10^{15}$ random numbers per second~\cite{misra_probabilistic_2022}.  Compare this to the reliability of transistor switching in conventional computers, where the probability of failure is less than $10^{-14}$~\cite{Koch1999}. It is unknown precisely how brains deal with this stochasticity, but its pervasiveness strongly suggests that the brain uses its own randomness as a computational resource rather than treating it as a defect that must be eliminated.  This suggests that a new class of parallel computing architectures could emerge from combining the computational principles of natural brains with physical sources of intrinsic randomness.  This would allow the natural stochasticity of electronic devices to play a part in large-scale parallel computations, relieving the burden imposed by requiring absolute reliability. 

Realizing the potential of probabilistic neural computation requires rethinking conventional parallel algorithms to incorporate stochastic elements from the bottom up. Additionally, techniques for controlling the randomness must be developed so that useful random numbers can be produced efficiently from the desired distributions. In this work, we propose neuromorphic circuits that demonstrate the capacity for intrinsic randomness to solve parallel computing problems and techniques for controlling device randomness to produce useful random numbers.

MAXCUT is a well known, NP-complete problem that has practical applications and serves as a model problem and testbed for both classical and beyond-Moore algorithm development~\cite{karp_reducibility_1972,GW95,khot_optimal_2007,basso_et_al:LIPIcs.TQC.2022.7}.  The problem requires partitioning the vertices of a graph into two disjoint classes such that the number of edges that span classes is maximized. MAXCUT has several stochastic approximation algorithms, which makes it an ideal target for developing new architectures leveraging large-scale parallel stochastic circuit elements for computational benefit.  

Stochastic approximation algorithms are compared via their approximation ratio, which is the ratio of the expected value of a stochastically generated solution to the maximum possible value. The stochastic approximation to MAXCUT with the largest known approximation ratio is the Goemans-Williamson algorithm~\cite{GW95}. The Goemans-Williamson algorithm provides the best approximation ratio achievable by any polynomial-time algorithm under the Unique Games Conjecture~\cite{khot_optimal_2007}. To generate solutions, this algorithm requires sampling from a Gaussian distribution with a specific covariance matrix obtained by solving a semi-definite program related to the adjacency matrix of the graph.  Our first neural circuit implements this sampling step by using simple neuron models to transform uniform device randomness into the required distribution.  This demonstrates the use of neuromorphic principles to transform an intrinsic source of randomness into a computationally useful distribution. 

Another stochastic approximation for MAXCUT is the Trevisan algorithm~\cite{trevisan_max_2012, soto_improved_2014}. Despite having a worse theoretical approximation ratio, in practice this algorithm generates solutions on par with the Goemans-Williamson algorithm~\cite{mirka_experimental_2022}.  To generate solutions, this algorithm requires computing the minimum eigenvector of the normalized adjacency matrix. Our second neuromorphic circuit implements this algorithm using the same circuit motif as above to generate random numbers with a specific correlation, but instead of sampling cuts from this distribution, we use these numbers to drive a synaptic plasticity rule (Oja's rule) inspired by the Hebbian principle in neuroscience~\cite{oja_principal_1992}.  This learning rule can be shown to converge to the desired eigenvector, from which the solution can be sampled.  This circuit solves the MAXCUT problem entirely within the circuit, without requiring any external preprocessing, demonstrating the capacity of neuromorphic circuits driven by intrinsic randomness to solve parallel computationally-relevant problems.  

Neuromorphic computing is having increasing impacts on non-cognitive problems relevant for parallel computing\cite{aimone_review_2022}. Unlike other hardware approaches to MAXCUT, our contributions directly instantiate state-of-the-art MAXCUT approximation algorithms on arbitrary graphs without requiring costly reconfiguration or conversion of the problem to an Ising model with pairwise interactions \cite{gyoten_area_2018, gyoten_enhancing_2018, yamaoka_20k-spin_2016}. Our use of hardware resources is scalable, requiring one neuron and one random device per vertex, and thus more efficient than parallel implementations of MAXCUT using GPUs \cite{cook_gpu-based_2019}. These properties make our contributions valuable to the expanding field of beyond-Moore parallel algorithms.

\section{MAXCUT Algorithms}
\subsection{The Goemans-Williamson MAXCUT Algorithm}

Given an $n$-vertex, $m$-edge graph $\mathcal{G} = (V, E)$ with vertex set $V$ and edge set $E$, the MAXCUT problem seeks a partition of the vertices into two disjoint subsets, $V = V_{-1} \cup V_{1}$ such that the number of edges that cross between the two subsets is maximized. By assigning either of the values $\{-1, 1\}$ to each vertex, the MAXCUT problem is equivalent to maximizing the function 
\begin{align*}
\max_{v} \quad & \frac{1}{2} \sum_{ij \in V} A_{ij}(1-v_i v_j) \\
\textrm{s.t.} \quad &v \in \{-1, 1\}^n.
\end{align*}
Here, $A_{ij}$ is the adjacency matrix of the graph $\mathcal{G}$. Let $\text{OPT}(\mathcal{G})$ be the maximum value of this function.

MAXCUT is known to be NP-complete \cite{karp_reducibility_1972}. Goemans and Williamson~\cite{GW95} described a relaxation of the above integer programming problem that yields an efficient approximation to MAXCUT with an approximation ratio of 0.878. The relaxation replaces the integer programming problem with a semidefinite programming problem given by
\begin{align*}
\max_w \quad & \frac{1}{2}\sum_{ij} A_{ij}(1 - w_i \cdot w_j)\\
\textrm{s.t.} \quad & w_i \in \mathcal{S}^{n-1},
\end{align*}
where $\mathcal{S}^{n-1}$ is the $(n-1)$-dimensional unit sphere in $\mathbb{R}^n$. Let $\text{SDP}(\mathcal{G})$ be value of the optimal solution of this semidefinite programming problem. Note that $\text{OPT}(\mathcal{G}) \leq \text{SDP}(\mathcal{G})$.

The solution of this semidefinite programming problem is a set of unit vectors $w_i$, one for each vertex in the graph. Given these vectors, a graph cut is generated by taking a random hyperplane through the origin and assigning the value $+1$ to vertices with vectors above the plane and $-1$ to vertices with vectors below the plane. 

One can see that the Goemans-Williamson algorithm has two steps: in the first step we solve an SDP, and in the second we round each unit vector $w_i$ to an integer $z_i \in \{-1, +1\}$, where $i \in V$. Bertsimas and Ye~\cite{BY98} observed that the rounding step can be implemented by sampling dependent standard normal random variables, with one variable per vertex. Specifically, suppose for each vertex $i$ we have a random variable $X_i$ following the standard normal distribution, and furthermore for each pair of vertices $i$ and $j$ the covariance between $X_i$ and $X_j$ is $w_i \cdot w_j$, where $w_i$ and $w_j$ are the unit vectors in the solution to the SDP. One can show that such a set of dependent random variables exists. Now define a (random) cut by assigning $+1$ to vertices $i$ where $X_i$ is positive and assigning $-1$ to vertices $i$ where $X_i$ is negative. One can show that the resulting cut has the same approximation guarantees as the cut returned by the Goemans-Williamson algorithm.
Hence in this paper we will sometimes refer to the rounding step as the sampling step.

\subsection{The Trevisan (Simple Spectral) Algorithm}

The Trevisan algorithm is another random approximation algorithm for MAXCUT \cite{trevisan_max_2012}.  Though it has a worse theoretical approximation ratio (0.631)\cite{soto_improved_2014} than the Goemans-Williamson algorithm, in practice it can perform just as well and has speed advantages~\cite{mirka_experimental_2022}. Here we consider a slight modification of the full Trevisan algorithm we refer to as the Trevisan Simple Spectral algorithm~\cite{mirka_experimental_2022}.

Given a graph $G = (V, E)$ with adjacency matrix $A$ and diagonal degree matrix $D$, we compute the normalized adjacency matrix $\mathcal{A} = D^{-1/2} A D^{1/2}$.  Next, the eigenvector corresponding to the minimum eigenvalue of the matrix $I + \mathcal{A}$ is computed.  The graph cut is obtained by thresholding the values of this eigenvector by sign. If $\textbf{u}$ is the minimum eigenvector of $\mathcal{A}$, then the graph cut is given by
$$
v_i = 
\begin{cases} 
      -1 & \textbf{u}_i \leq 0 \\
      1 & \textbf{u}_i > 0 
\end{cases}
$$

\section{Neuromorphic Concepts}
\subsection{Stochastic Devices}

Physical microelectronics display intrinsic stochasticity due the physics behind their operation. Typically this stochasticity is observed as random switching between two or more states. While normally a nuisance, the details of this stochastic behavior are under active research to develop devices with tunable statistics for probabilistic computing applications \cite{misra_probabilistic_2022, rehm_stochastic_2022, bernardo-gavito_extracting_2017}. In our work, we idealize stochastic devices as analogous to ``coin flips'' such that at any given time, the device can be in one of two states (``heads'' or ``tails''; ``0'' or ``1'') with a specific probability.  In our circuits, we assume random devices behave as fair coins. That is, each state has a probability of 0.5. Thus, a random device is modeled as a source for a random bit stream with equal probabilities of 0 or 1. Magnetic tunnel junctions~\cite{kent_new_2015, rehm_stochastic_2022} and tunnel diodes~\cite{bernardo-gavito_extracting_2017} are examples of two classes of devices actively being developed to meet these requirements.  

\subsection{Leaky Integrate and Fire Neurons}

The leaky integrate and fire (LIF) neuron is a simplified model of biological neurons, readily implemented in hardware~\cite{indiveri_neuromorphic_2011}, that captures a biological neuron's capacity for temporal integration of synaptic inputs along with discontinuous spiking. The model integrates synaptic currents with a membrane capacitance into a membrane potential that is continuously discharged by a leak conductance.  When the integrated membrane potential reaches some threshold, a spike is emitted and the membrane potential is reset to some defined value.  In between spike events, the membrane potential evolves according to the differential equation
\[
C\frac{dV}{dt} = -\frac{V}{R} + I_{\textrm{tot}}.
\]
Here, $V$ is the membrane potential, $C$ and $R$ are the membrane capacitance and leak resistance, respectively, and $I_{\textrm{tot}}$ is the total synaptic input current. 

When a single LIF neuron receives large numbers of stochastic input currents, the membrane potential approximates a one-dimensional random walk~\cite{Koch1999}.  The leak conductance stabilizes this walk around an analytically computable mean
\[
\langle V \rangle = R\langle I_{\textrm{tot}} \rangle
\]
and variance
\[
\textrm{Var}(V) = \frac{R}{C} \textrm{Var}(I_{\textrm{tot}}).
\]

\subsection{LIF Covariances}
For a population of $n$ LIF neurons integrating random binary inputs generated by $r$ random devices, the expression for the membrane potential dynamics of a single LIF neuron becomes
\[
C\frac{dV_i}{dt} = -\frac{V_i}{R} + \sum_{\alpha} W_{i\alpha} s_{\alpha}.
\]
Here $W_{i\alpha}$ is the real-valued connection weight between device $\alpha$ and LIF neuron $i$. The variable $s_{\alpha}$ is the state of device $\alpha$ and takes the values $\{0, 1\}$.

Shared or inverted input between two LIF neurons induces correlations or anticorrelations in their membrane potentials, respectively.  The expression for the covariance between the membrane potentials of neurons $i$ and $j$ is
\[
\textrm{Cov}(V_i, V_j) = \frac{R}{C}\sum_{\alpha \beta} W_{i \alpha} W_{j \beta} \textrm{Cov}(s_{\alpha}, s_{\beta}).
\]

In other words, the LIF membrane covariances are a linear transformation of the covariances of the random device pool. The device covariance matrix defines an inner product on the space of weight vectors for each LIF neuron.  If the devices are independent, then the device covariance matrix is diagonal.  Thus, the LIF neuron population transforms the device randomness into a set of Gaussian processes with covariance proportional to the Gram matrix of the weight vectors. In what follows, choosing the weights appropriately allows this circuit motif to supply random samples with the appropriate covariances for the stochastic MAXCUT approximation algorithms.

\begin{figure}
    \centering
    \includegraphics[width=0.45\textwidth]{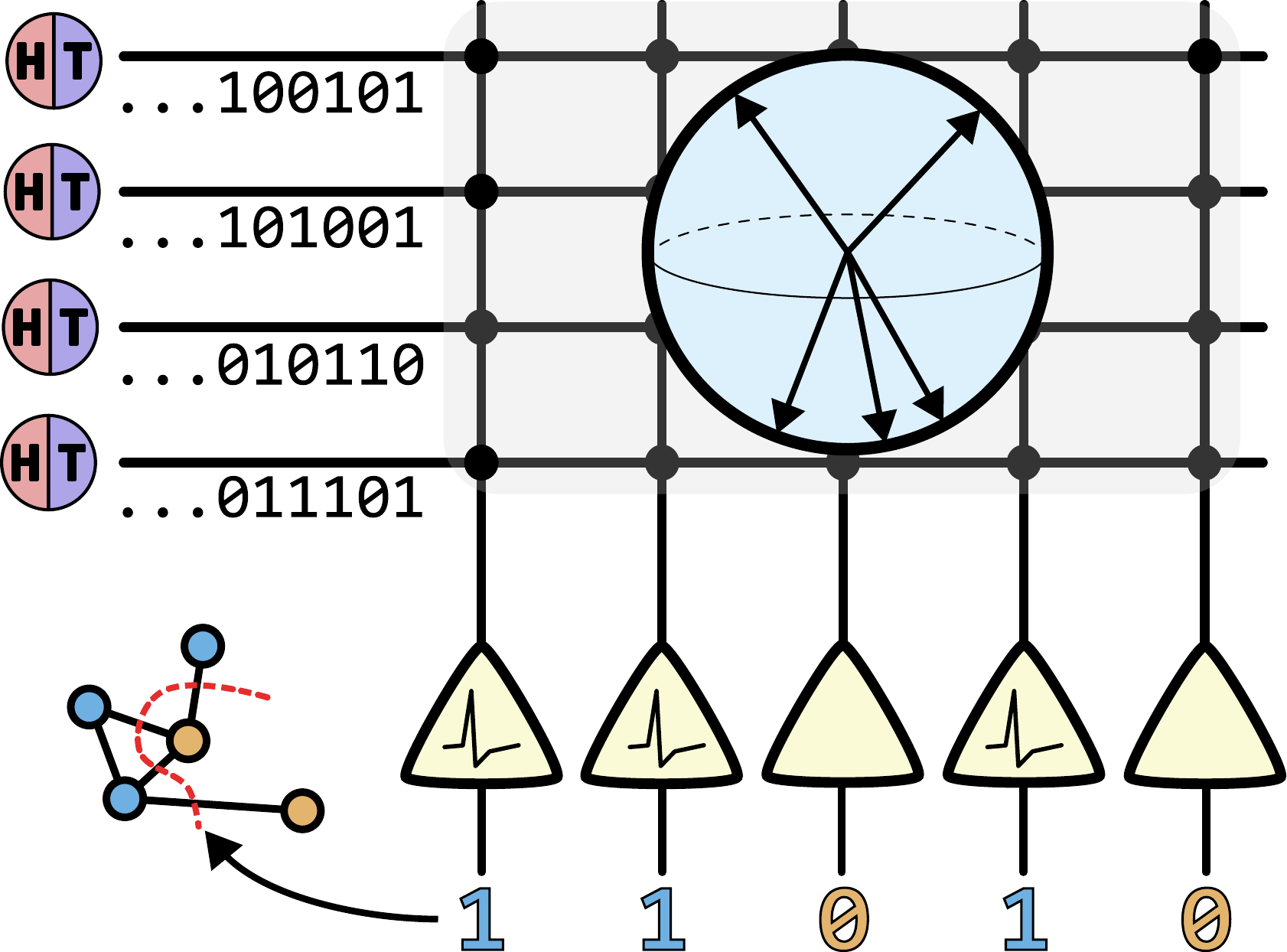}
    \caption{LIF-Goemans-Williamson circuit implementing the sampling stage of the Goemans-Williamson algorithm.  Spikes from the LIF population correspond to binary labels on the vertices of the graph, defining the cut.  The covariances of the LIF membrane potentials are determined by the weight matrix from the random devices (left) to the LIF population.  Each LIF neuron's weight vector is set proportional to a vector determined through the solution to the Goemans-Williamson SDP.}
    \label{fig:LIF_GW}
\end{figure}

\subsection{Synaptic Plasticity: Oja's Rule}

In neuroscience, the guiding principle of synaptic plasticity is captured by the adage ``neurons that fire together, wire together.''  This is the Hebbian learning principle~\cite{hebb_organization_1949}. If $\textbf{w}$ is the weight vector between presynaptic neuron activity $\textbf{x}$ and postsynaptic neuron activity $y$, the simplest instantiation of this principle is given the formula
$$
\Delta \textbf{w} = y\textbf{x}
$$
As stated, this rule is unstable.  Oja presented a modification to this plasticity rule that preserved the Hebbian principle but enforced weight stability~\cite{oja_simplified_1982}.  Oja's rule is given by the formula
$$
\Delta \textbf{w} = y(\textbf{x} - y\textbf{w})
$$

Oja proved that under mild assumptions this rule forces the weight vector to converge to the first principle component of the covariance matrix of the inputs, or equivalently the eigenvector corresponding the the largest eigenvalue. 

By considering anti-Hebbian plasticity, Oja derived a related, stabilized learning rule that converges to the \textit{minimum} eigenvector of the covariance matrix~\cite{oja_principal_1992}
$$
\Delta \textbf{w} = -y\textbf{x} +(y^2 +1 - \textbf{w}^T \textbf{w})\textbf{w}
$$

By providing inputs with covariance proportional to the adjacency matrix of the graph as used in Trevisan's algorithm, Oja's anti-Hebbian rule can find the minimum eigenvector of this matrix, yielding an approximate solution to MAXCUT.

\section{Circuits}

\subsection{LIF-Goemans-Williamson}

Figure~\ref{fig:LIF_GW} shows a neural circuit that implements the sampling step of the Goemans-Williamson algorithm.  The requirement is to generate binarized samples from a Gaussian distribution with specified covariance matrix $C$.  We refer to this circuit with the abbreviation LIF-GW.

For a graph $G$, the Goemans-Williamson SDP is solved to yield a set of $n$ unit vectors in $r$ dimensions, where $r$ is the rank of the solution and $n$ is the number of vertices.  These vectors can be combined into the $n$ by $r$ dimensional matrix $W_{GW}$.

The circuit consists of a pool of $r$ random devices connected to $n$ LIF neurons.  The synaptic weights between the devices and the neurons are chosen proportional to the corresponding entries in $W_{GW}$.  The precise magnitudes of these weights are not critical; what matter are their relative values, as these ratios determine the LIF covariances.  This allows the circuit to be adapted to specific hardware implementations imposing constraints on the range of available weights. For our tests, we used a fixed rank of 4 for all graphs.

Choosing the weights proportional to the solution to the SDP yields membrane covariances proportional to those required by the Goemans-Williamson algorithm.  The spiking threshold of the LIF neurons implements a rounding and sampling operation that we map to graph cuts.  Neurons that spike together on a given timestep map to vertices on one side of the cut, and neurons that are silent on a given timestep map to vertices on the other side of the cut. 

\subsection{LIF-Trevisan}
The second neural circuit (Figure \ref{fig:LIF_TR}) implements the simple spectral modification of Trevisan's algorithm \cite{trevisan_max_2012, mirka_experimental_2022} and we refer to it as either LIF-Trevisan or LIF-TR.  Like the LIF-GW circuit, the first stage consists of a population of LIF neurons, one for each vertex in the graph, driven by a pool of random devices.  Next, the output of the LIF population is fed onto a single LIF neuron.  The output of this Stage-2 neuron is discarded; what matters is the weight vector $w$ linking the Stage-1 LIF population to Stage-2.  This weight vector is controlled by Oja's anti-Hebbian plasticity rule.  This forces the weight vector $w$ to converge onto the minimum eigenvector of the LIF covariance matrix.  

The LIF covariance matrix is determined by the connection weights between the random devices and the LIF population.  These are set proportional to the Trevisan matrix, which is the sum $I + D^{-1/2}AD^{-1/2}$ of the identity plus the normalized adjacency matrix of the graph.  In this way, the LIF-Trevisan circuit does not require solving an SDP offline.  

\begin{figure}
    \centering
    \includegraphics[width=0.45\textwidth]{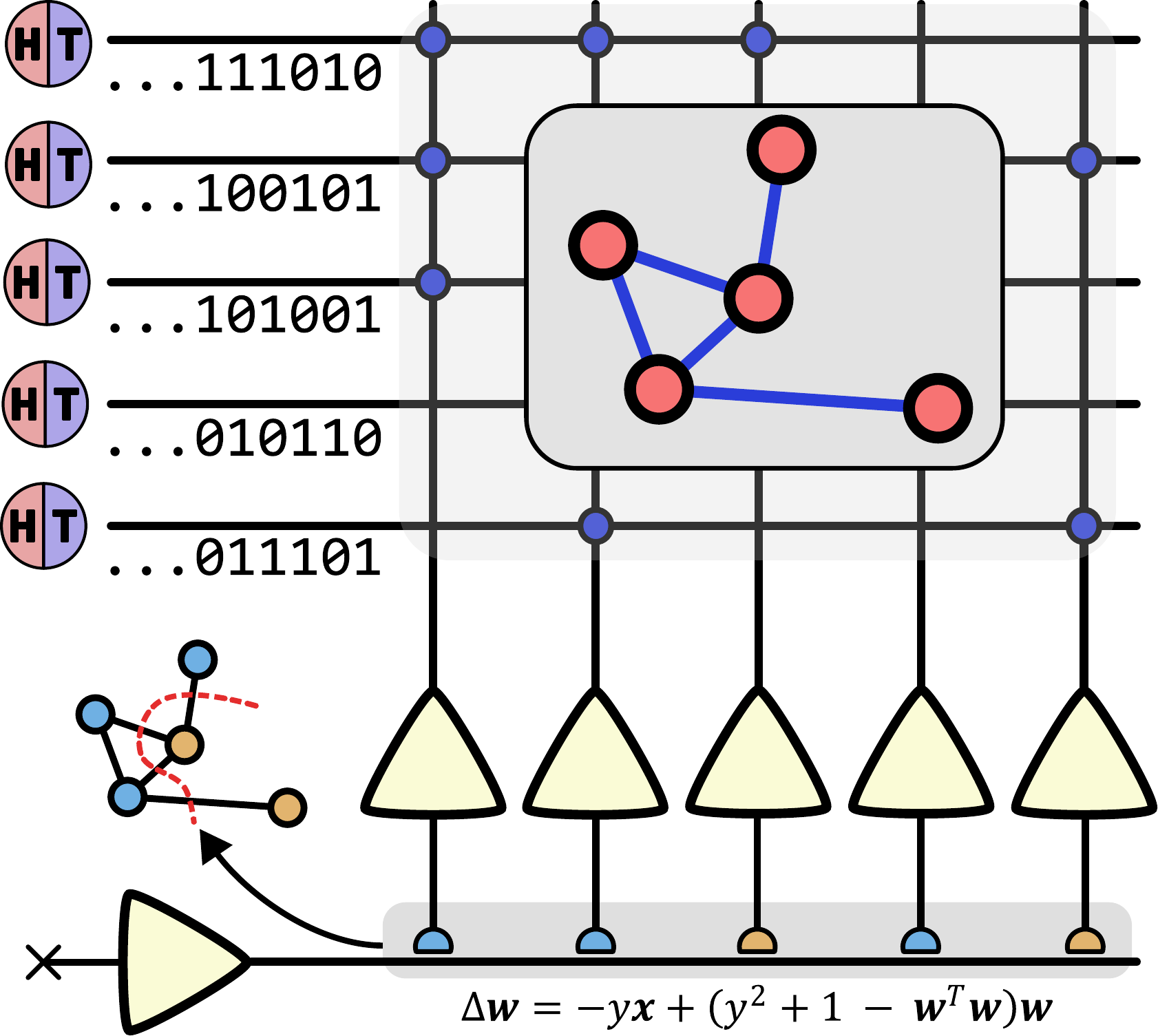}
    \caption{LIF-Trevisan circuit implementing a stochastic approximation to MAXCUT by combining hardware randomness with anti-Hebbian synaptic plasticity. The connection weights between the random device pool (left) and the LIF neurons are set proportional to the adjacency matrix of the graph.  The activity of the LIF neurons drives synaptic plasticity on the weight vector onto an output neuron.  The solution is sampled by thresholding this weight vector by sign: excitatory, positive weights correspond to one side of the cut and inhibitory, negative weights correspond to the other side. The output of the output neuron is ignored. }
    \label{fig:LIF_TR}
\end{figure}

\section{Results}

\begin{figure*}
    \centering
    \includegraphics[width=1\textwidth]{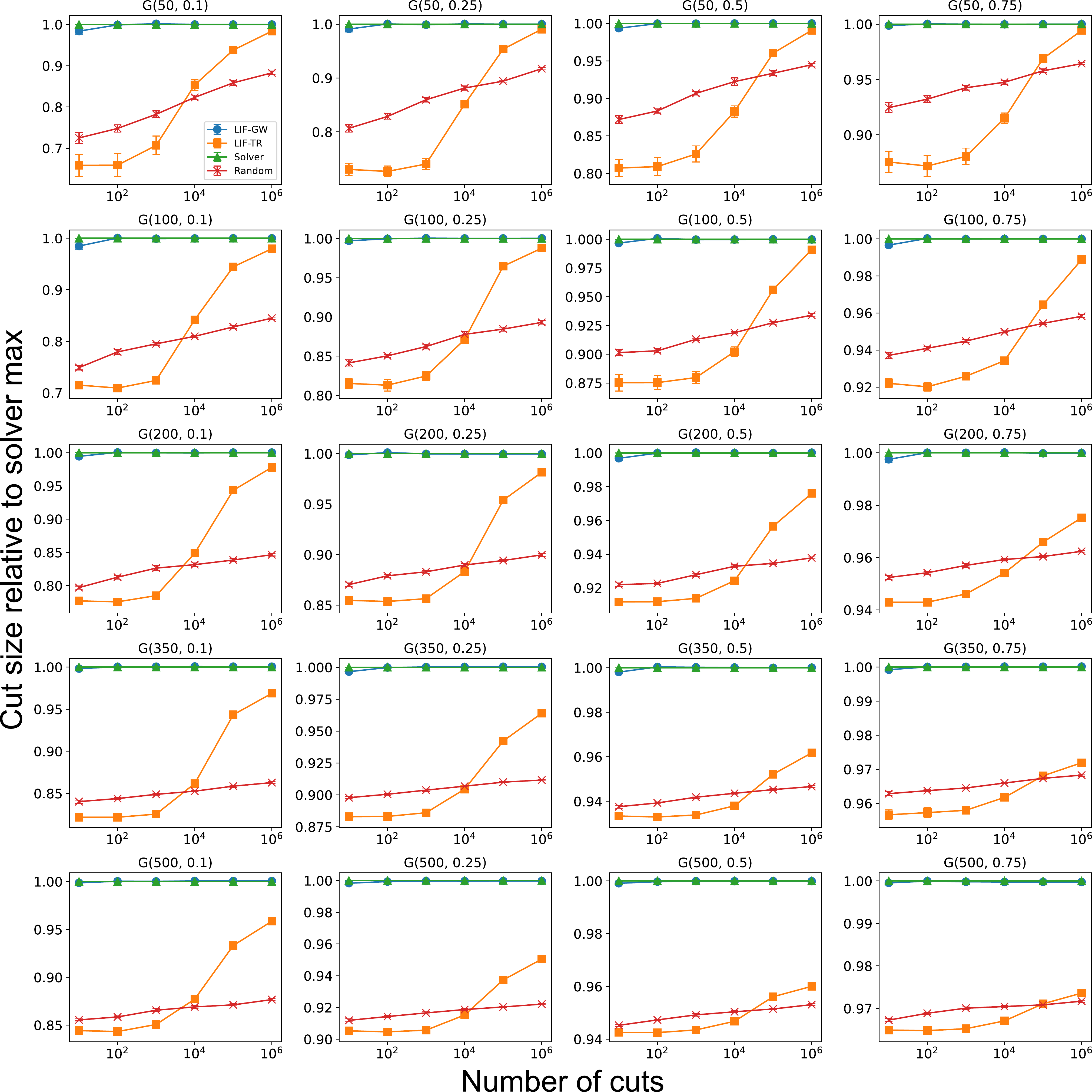}
    \caption{Maximum cut weight relative to software Goemans-Williamson solver (green triangle curve) as a function of the number of samples for Erd\H{o}s-R\'{e}nyi random graphs. Rows correspond to fixed numbers of vertices $n$ and columns correspond to fixed connection probabilities $p$. Panel title gives graph parameters $G(n, p)$. Error bars correspond to standard error of the mean over 10 independently generated graphs from each graph class. Blue circles: LIF-GW circuit; orange squares: LIF-TR circuit; green triangles: software solver; red X's: random graph cuts. Blue and green curves overlap.}
    \label{fig:circuit_results}
\end{figure*}

We simulated these circuits and quantified their ability to generate graph cuts.  Following~\cite{mirka_experimental_2022} we evaluated the circuits on Erd\H{o}s-R\'{e}nyi random graphs with a number of vertices $n$ in $\{50, 100, 200, 350, 500\}$ and a connection probability $p$ in $\{0.1, 0.25, 0.5, 0.75\}$. We generated 10 distinct random graphs per $(n, p)$ combination, yielding 200 total graphs. We generated $2^{20}$ graph cuts per circuit, per graph.  We compared the circuit-generated cut weights to cut weights generated by a generic SDP solver (PyManOpt~\cite{pymanopt}) and cut weights generated by a purely random assignment of vertices to sides of a cut. As described, circuits were driven by a simulated pool of random devices.  Each device was assumed to have two states, and have a probability of 0.5 of being in any given state at each time step.  

Figure \ref{fig:circuit_results} shows that, as expected, the LIF-GW circuit matches the performance of the generic solver.  This also validates the proposed circuit motif using LIF neurons to translate hardware randomness into Gaussian processes with desired covariances.
The LIF-Trevisan circuit shows performance that increases over time, approaching the performance of the solver, due to the on-line learning of the solution through Oja's rule.  In all cases, the LIF-Trevisan circuit eventually outperforms the random algorithm. The trajectory of the LIF-Trevisan circuit's performance suggests that the rate of convergence of the plasticity depends on the graph parameters, but there is no indication that the performance of this circuit is saturated within the number of samples considered here.

We next evaluated our circuits on empirical graphs taken from the Network Repository~\cite{nr}. We picked the same graphs tested in~\cite{mirka_experimental_2022}.  Figure~\ref{fig:circuit_results_nrvis} shows the performance of our circuits compared to SDP-solver and random cuts.  Consistent with out results on Erd\H{o}s-R\'{e}nyi graphs, we found that the LIF-GW sampling circuit matched the performance of the software solver.  We found that the LIF-Trevisan circuit was able to outperform randomly-generated cuts and, occasionally, exactly match the solver-generated cuts, after evolving the circuit with synaptic plasticity for sufficiently many samples.  This is consistent with the results of \cite{mirka_experimental_2022}, who found that in some cases the simplified approximation algorithms to MAXCUT matched or outperformed cuts generated by the full Goemans-Williamson algorithm on empirical graphs, even though the simplified algorithms have worse approximation guarantees.  The maximum cut values for each circuit for each graph are presented in Table \ref{tab:maxcuts}, which are in agreement with the maximum cut sizes found in~\cite{mirka_experimental_2022} (rightmost column).

\begin{figure*}
    \centering
    \includegraphics[width=1\textwidth]{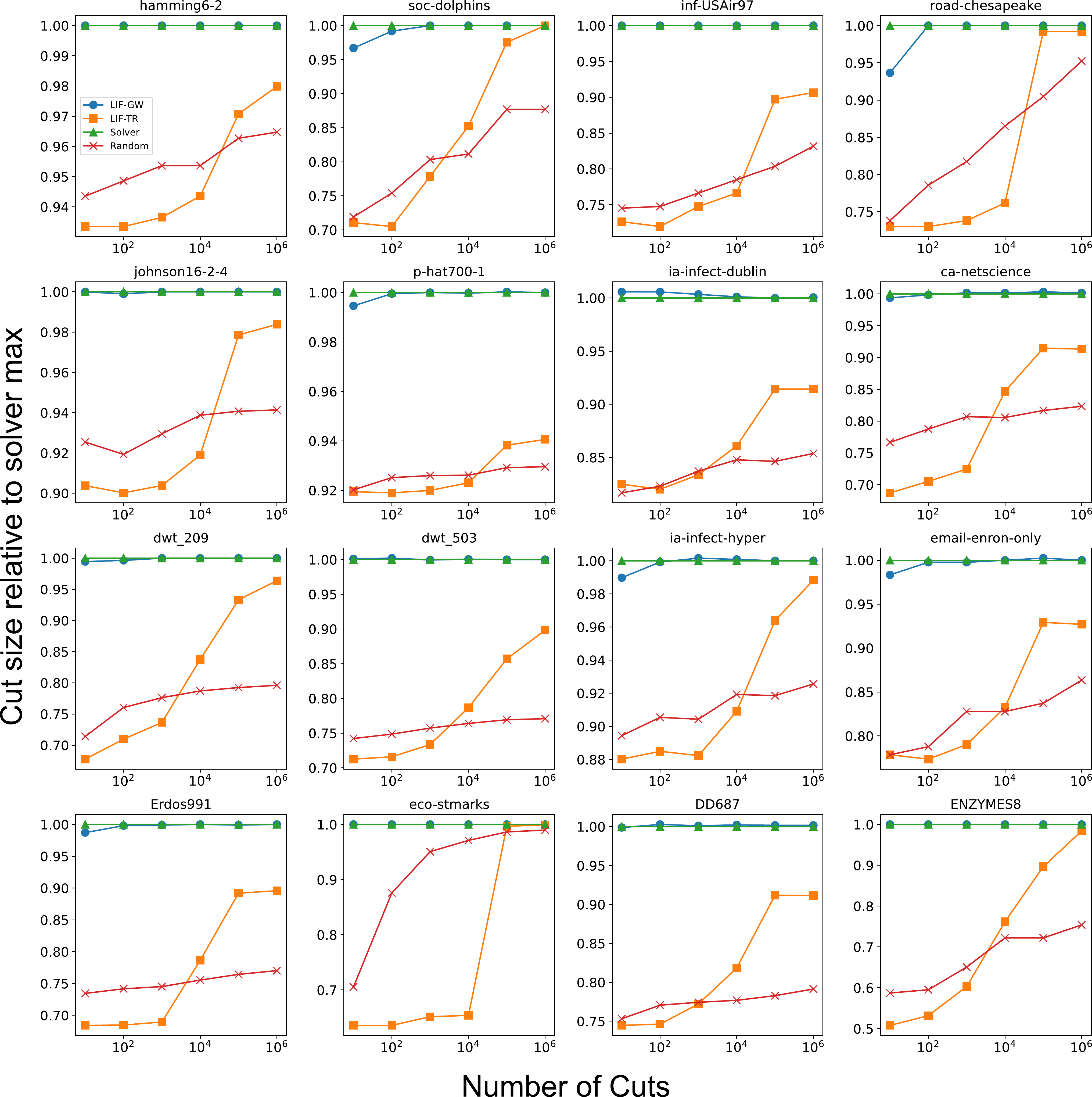}
    \caption{Maximum cut relative to solver as a function of the number of samples for empirical graphs taken from the Network Repository. Each panel represents a single graph, thus there are no error bars. Panel title identifies the graph dataset. Blue circles: LIF-GW circuit; orange squares: LIF-TR circuit; green triangles: software solver; red X's: random graph cuts. Blue and green curves overlap.}
    \label{fig:circuit_results_nrvis}
\end{figure*}

\begin{table*}[h!]
\centering
 \caption{Maximum cut values for the different neuromorphic circuits applied to empirical graphs from the Network Repository \cite{nr}.}
 \label{tab:maxcuts}
 \begin{tabular}{|c c c c c c |} 
 \hline
Graph & LIF-GW & LIF-TR & Solver & Random & \cite{mirka_experimental_2022}\\ [0.5ex] 
 \hline\hline
hamming6-2 & 992 & 972 & 992 & 957 & 992  \\ 
soc-dolphins & 122 & 122 & 122 & 107 & 121  \\ 
inf-USAir97 & 107 & 97 & 107 & 89 & 107 \\ 
road-chesapeake & 126 & 125 & 126 & 120 & 125 \\ 
johnson16-2-4 & 3036 & 2987 & 3036 & 2858 & 3036 \\ 
p-hat700-1 & 33350 & 31369 & 33351 & 31002 & 33050 \\ 
ia-infect-dublin & 1751 & 1600 & 1750 & 1494 & 1664  \\ 
ca-netscience & 635 & 579 & 634 & 522 & 611  \\ 
dwt-209 & 554 & 534 & 554 & 441 & 540 \\ 
dwt-503 & 1937 & 1740 & 1937 & 1493 & 1921 \\ 
ia-infect-hyper & 1277 & 1262 & 1277 & 1182 & 1233  \\ 
email-enron-only & 425 & 394 & 425 & 367 & 413 \\ 
Erdos991 & 1027 & 920 & 1027 & 791 & 934  \\ 
eco-stmarks & 1765 & 1764 & 1765 & 1747 & 1190  \\ 
DD687 & 1786 & 1625 & 1783 & 1411 & 1680\\ 
ENZYMES8 & 126 & 124 & 126 & 95 & 126 \\ 
 \hline
 \end{tabular}

\end{table*}

\section{Discussion}

We have presented two neuromorphic circuits that transform the activity of a pool of random devices into useful distributions that solve a computational problem, in this case, MAXCUT. Our approach combines insights from theoretical computer science, neuroscience, and materials science to show that probabilistic neural computing is a viable path to new computational architectures. Our circuits display competitive performance with traditional software solvers. Consistent with prior work~\cite{mirka_experimental_2022}, we find that the simple spectral Trevisan algorithm performs in practice nearly as well as the gold-standard Goemans-Williamson algorithm. Our results for the Trevisan circuit suggest that its performance can be expected to improve beyond the $2^{20}$ samples considered here.  While in this work we considered only a single example problem, MAXCUT is a special case of a larger class of problems known as constraint satisfaction problems, which include problems like maximum directed cut (MAXDICUT) and maximum 2-satisfiability (MAX2SAT). As such, our circuits may extend to more general probabilistic neural approaches for solving discrete optimization problems. For instance, using results due to Goemans and Williamson~\cite{GW95}, our LIF-GW circuit can implement sampling steps for algorithms for MAXDICUT and MAX2SAT that yield approximation ratios of 0.796 and 0.878, respectively.

Our circuits present a trade-off in neuromorphic implementations of combinatorial optimization. The LIF-GW circuit requires fewer random devices and delivers superb solutions rapidly, but requires a substantial commitment of offline resources (i.e. solving a semi-definite program) to initialize.  Conversely, the LIF-Trevisan circuit requires as many random devices as vertices in the graph, and takes many more samples to reach comparable performance.  However, this circuit avoids offline computations, which are a significant fraction of the running time of these algorithms~\cite{mirka_experimental_2022}. This prior work also used far fewer samples (100) to compare the algorithms than our $2^{20}$. While the number of samples required suggests a disadvantage, at the speed of hardware, the greater number of samples required will likely be a trivial increase in the running time compared to a software implementation. Current hardware implementations of LIF neurons operate with time constants on the order of 1 nanosecond~\cite{hassan_magnetic_2018, brigner_shape-based_2019}.  Using this value as a reference time step for a hardware implementation of these circuits, the circuits could generate millions of samples in the time required for a software simple spectral computation ($\sim 10 \textrm{ms}$), or billions of samples in the time required to solve and sample the Goemans-Williams SDP~\cite{mirka_experimental_2022}. The convergence rate of the synaptic plasticity in our LIF-TR circuit depends on both the circuit parameters and graph structure in complicated ways, but this dependence could be formalized or optimized in future work. Furthermore, the requirement for greater numbers of random devices is likely not a limitation, as the current trajectory for implementing the stochastic devices required for these circuits shows promising scaling advantages~\cite{misra_probabilistic_2022}.

Our simulations model random devices as perfectly fair coins generating random, independent bit streams.  These assumptions are necessarily approximations to the true behavior of a random device, which may display the statistics of an unfair coin, show internal or external correlations, or display statistics that drift over time.  These imperfections might have an impact on the performance of the circuits presented.  The key circuit motif in each circuit implements the central limit theorem through the integration of large numbers of random devices.  Thus, we expect robustness to deviations of individual devices from the idealized perfect coin as the number of devices grows. While there is a growing realization that stochastic devices can provide robust random bit streams\cite{rehm_stochastic_2022}, there currently are few standards for what makes a good true random number generator for randomized algorithms. For this reason, the circuits described here provide a much needed benchmark for device physicists to incorporate physically-detailed device models to assess the impact of device variability. 

Neuromorphic computing is having a growing impact on graph algorithms \cite{aimone_review_2022}. Previous work has found neuromorphic solutions to graph problems such as max flow \cite{kay_neuromorphic_2021}, cycle detection\cite{kay_neuromorphic_2021}, shortest paths\cite{kay_neuromorphic_2020, aimone_dynamic_2019, aimone_provable_2021}, and spanning trees \cite{kay_neuromorphic_2020}.  This prior work has exploited connections between the graph structure of neural networks and the corresponding graph problems. In contrast, our work uses the statistical behavior of neural circuits and learning through synaptic plasticity to solve a new class of graph problems. Incorporating learning into neuromorphic circuits to solve specific computational problems is comparatively under-explored, and thus our work expands the neuromorphic acceleration of graph algorithms in a new direction. 

Our approach presents a different strategy for incorporating neuroscientific insight into parallel computation.  The most successful application of neuroscience principles to date occurs in the field of deep learning, which is based on connectionist principles inherited from early neuroanatomical studies. In contrast, our circuits use the integrative, statistical properties of neurons to achieve different kinds of computations.  This was informed by early theoretical developments, like Oja's rule for synaptic plasticity~\cite{oja_principal_1992, oja_simplified_1982}.  These developments have been well-known for decades, but have not had as strong an influence on computation. Similar to how backpropagation was known for many years before physical implementations achieved the scale necessary to reveal its utility, we expect that recent advances in stochastic devices and neuromorphic hardware will reveal the utility of probabilistic neural computation.

\section*{Acknowledgment}

The authors acknowledge financial support from the DOE Office of Science (ASCR / BES) for our Microelectronics Co-Design project COINFLIPS. Sandia National Laboratories is a multi-mission laboratory managed and operated by National Technology and Engineering Solutions of Sandia, LLC, a wholly owned subsidiary of Honeywell International, Inc., for the U.S. Department of Energy's National Nuclear Security Administration under contract DE-NA0003525. This paper describes technical results and analysis. Any subjective views or opinions that might be expressed in the paper do not necessarily represent the views of the U.S. Department of Energy or the United States Government. SAND Number: SAND2022-13654 O

\bibliographystyle{plain}
\bibliography{main}
\end{document}